\documentclass[conference]{IEEEtran}
\IEEEoverridecommandlockouts
\usepackage{cite}
\usepackage{amsmath,amssymb,amsfonts}
\usepackage{algorithmic}
\usepackage{graphicx}
\usepackage{textcomp}
\usepackage{xcolor}
\usepackage{gensymb}
\usepackage{multirow}

\def\BibTeX{{\rm B\kern-.05em{\sc i\kern-.025em b}\kern-.08em
    T\kern-.1667em\lower.7ex\hbox{E}\kern-.125emX}}
\begin{document}

\title{End-to-end system for object detection from sub-sampled radar data
}

\author{Madhumitha Sakthi, Ahmed Tewfik, Marius Arvinte, Haris Vikalo \thanks{The authors are with the Department of Electrical and Computer Engineering, The University of Texas at Austin. E-mails: {madhumithasakthi.iyer@utexas.edu, tewfik@austin.utexas.edu, arvinte@utexas.edu, hvikalo@ece.utexas.edu.} code:https://github.com/Madhusakth/RADIATE-Adaptive-CS}}


\maketitle

\begin{abstract}
Robust and accurate sensing is of critical importance for advancing autonomous automotive systems. The need to acquire situational awareness in complex urban conditions using sensors such as radar has motivated research on power and latency-efficient signal acquisition methods. In this paper, we present an end-to-end signal processing pipeline, capable of operating in extreme weather conditions, that relies on sub-sampled radar data to perform object detection in vehicular settings. The results of the object detection are further utilized to sub-sample forthcoming radar data, which stands in contrast to prior work where the sub-sampling relies on image information. We show robust detection based on radar data reconstructed using 20\% of samples under extreme weather conditions such as snow or fog, and on low-illuminated nights. Additionally, we generate 20\% sampled radar data in a fine-tuning set and show 1.1\% gain in AP50 across scenes and 3\% AP50 gain in motorway condition. 
\end{abstract}


\section{Introduction}
A thorough understanding of surroundings is vital for the safety of autonomous driving systems. Similar to humans who while driving rely on multiple sensor information such as sound and vision, these systems acquire data from a variety of sensors including e.g. image, radar and LIDAR. Among those, radar was demonstrated to enable accurate object detection whether used in conjunction with other sensing modalities (e.g., images) \cite{centre-fusion,radar1,radar2,rrpn} or alone \cite{ROD-Net, track-detect-radar, efficient-ROD, ROD-squeeze-excite, radiate-object-detection, ROD-dimension-apart-nw, ana-radar-object-detection}. The ability to achieve accurate object detection without relying upon data other than radar is critical in extreme weather conditions such as snow, fog or rain where the image sensors may struggle to provide the information needed to develop situational awareness \cite{rad-cam-fusion}.  

When a sensor rapidly collects information about a vehicle's environment, it is necessary to reduce the data rate while maintaining the quality of acquired information. To this end, signal acquisition often relies on compressed sensing (CS) to collect data at a sub-Nyquist rate without compromising the quality of information \cite{cs-candes}. The CS algorithms typically exhibit a trade-off between signal reconstruction quality and sampling rate. In \cite{CS-SAR-Radar}, the CS framework is utilized to acquire Synthetic Aperture Radar (SAR) data and achieve robust reconstruction with 70\% of the samples. In \cite{radar-cs-40}, authors demonstrate efficient reconstruction of a frequency-modulated continuous wave (FMCW) radar using 40\% of the samples. A CS-based signal acquisition methodology for noise radar with a 30\% sampling rate was presented in \cite{radar-cs2}. The authors of \cite{radar-cs3} compare the performance of Orthogonal Matching Pursuit (OMP) and Basis Pursuit De-noising (BPDN) in applications to the direction-of-arrival estimation problem; note that while OMP achieves more accurate reconstruction, it generally requires more measurements than basis pursuit \cite{OmpVsBp}. This motivates the use of the basis pursuit (BP) algorithm in our current work.

\begin{figure*}[ht!]
\begin{center}
\includegraphics[width=15.5cm]{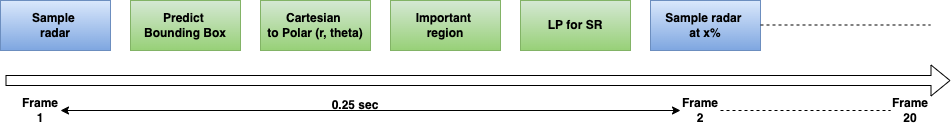}
\centering
\caption{\label{Algorithm}The overall sampling pipeline for every 20 frames across scenes. }
\label{Algo}
\end{center}
\vspace{-6mm}
\end{figure*}

Adaptive CS is a technique used to increase the sampling rate for important signal regions \cite{adaptive-radar-cs1}, \cite{adaptive-radar-cs2},  \cite{adaptive-radar-cs3}. In \cite{adaptive-radar-cs1}, the authors present a pulsed radar undersampled acquisition method that utilizes the previously received pulse interval and applies a constant false alarm rate (CFAR) detection technique to determine the importance coefficients for the present interval. 
In \cite{adaptive-radar-cs2}, adaptive CS is utilized in static settings to improve target tracking performance. In contrast, in the current paper for an autonomous vehicle, while both the vehicle and objects are potentially moving, we use adaptive CS algorithm for radar acquisition. 
Finally, \cite{adaptive-radar-cs3} presents an adaptive CS algorithm which aims to optimize the measurement matrix in a setting where only the targets are moving, leading to increased performance but, this also increased the computational complexity of the algorithm. In the proposed approach, the measurement matrix size is increased for certain regions of a radar frame using linear programming (LP) problem formulation; this allocates a larger sampling budget to important regions while keeping the overall sampling budget and reconstruction complexity under control.

In this paper, we develop an efficient signal acquisition/processing pipeline for radar-based object detection that uses the detection result from the current frame to sub-sample the subsequent radar frame via adaptive compressed sensing. In particular, we build upon \cite{cs-object} where the results of image-based object detection were used to identify the important regions in radar. To accommodate all weather conditions, the approach in the current paper removes the need for image data and instead relies on radar-based object detection results for the subsequent radar frame. 
We test our sub-sampled radar on an object detection task and show performance comparable to the fully-sampled radar. In particular, we test the method on the RADIATE dataset \cite{radiate} collected in extreme weather conditions and demonstrate robust detection based on radar data acquired using only 20\% of the samples. 
Finally, since task-based fine-tuning is known to improve the network performance, we generated 20\% sampled radar data and used the sub-sampled radar frames to fine-tune the object detection network. This resulted in 1.1\% gain in AP50 across all the scenes and 3\% gain in AP50 in the motorway condition. 

\section{Method}

\subsection{Data}
The RADIATE dataset \cite{radiate} was collected in multiple extreme weather conditions, and consists of radar, lidar, camera, GPS sensor data. The radar data was collected by the Navtech CTS350-X with 360\degree Horizontal Field of View (HFoV) and 100-meter range at 4Hz, resulting in a range-azimuth image of size 400x576 where the rows represent the angle and the column represent the range. The set contains 300 hours of annotated radar data. 
In addition to the sensor data, the authors released object annotation on the Cartesian radar images. Although the released annotations are fine-grained, similar to the \cite{radiate}, we classify objects as vehicle or background by defining the vehicle class as either car, bus, bicycle, truck, van or motorbike. We selected about 20 frames for each night, snow, foggy, motorway and city scene conditions, and used them to test our acquisition algorithm. In the fine-tuning case, we generated 200 20\% sampled radar data based on the previous frame's object detection result. 

\subsection{Adaptive radar sampling}
\begin{figure}[ht!]
\begin{center}
\includegraphics[width=8cm]{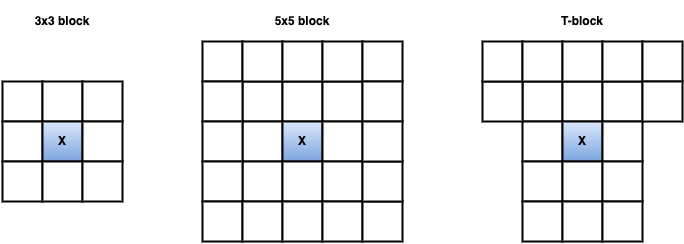}
\centering
\caption{\label{sampling-pattern}The 3x3 sampling pattern for small objects, 5x5 sampling pattern for big object and T-shaped pattern for objects beyond 50m from the AV. }
\label{Algo}
\end{center}
\end{figure}

In adaptive CS, the measurement region is split into blocks and the number of measurements allocated to each block varies according to a parameter. In each of the blocks, standard CS is performed by collecting $m$ measurements. To ensure robust reconstruction using the compressed measurements, the following assumptions are made. First, the measured signal is sparse in some domains. Specifically, we assume that the signal is sparse in the Discrete Cosine Transform (DCT) domain. Second, the measurement matrix exhibits the restricted isometry property \cite{Binary-CS-Image}. In each block, given the original signal $x \in R^n$ acquired using the random measurement matrix $\phi \in R^{mxn}$, we obtain measurements $y \in R^m$. The original signal $x$ is recovered using the basis pursuit (BP) algorithm as $\min_{x} \|\theta x\|_1$  $s.t.$  $\phi x = y $ with $\theta$ as the DCT transformation matrix \cite{image-adaptive-cs}.

In the RADIATE dataset, the radar data is captured at 4Hz, i.e., every 0.25 seconds. The radar frame in the polar domain of size $400 \times 576$ is split into $20 \times 48$ sized blocks; this amounts to 18\degree and 8.4 meters in range. In the baseline acquisition algorithm (Standard-CS) that deploys a uniform sampling rate, all the blocks were acquired using the same sampling rate of 10\%, 20\% or 30\%. 
That is, for each  $20 \times 48$ sized block, in the case of 10\% sampling rate, 96 samples were acquired using the measurement matrix and this information was used to reconstruct the original block using the BP algorithm. Similarly, for a radar frame of size $400 \times 576$, all 240 blocks were sampled and reconstructed to form the 10\% uniform sampled radar data. 
In the proposed algorithm, as shown in Figure \ref{Algorithm} the first frame is fully sampled and processed by the object detection network. However, the acquisition of radar data takes place in the polar domain. Specifically, in case of Navtech hardware, 400 measurements are acquired in each cycle, resulting in a 360\degree HFoV and 100 meters in range data. The polar data is converted to the Cartesian domain which leads to bird-eye view radar images. In \cite{radiate}, object annotations are provided in the Cartesian domain and thus the network is trained to perform object detection by processing this Cartesian radar data. Once the bounding boxes are predicted for a given frame, the object's center is converted to polar coordinates which are then used to identify the block in which the object is present. Since the next frame is acquired in 0.25 seconds, the region surrounding the current radar block is also marked important in order to account for the motion of the detected object as well as the motion of the Autonomous Vehicle (AV). 

Each polar block covers about 8.4 meters in length; this is sufficient to capture the moving vehicle within the adjacent block while the object moves at most 80 miles per hour. Hence, in Rad-info-1, if the detected object is small -- for example, a car or motorbike -- 
then the $3 \times 3$ block sampling pattern shown in Figure \ref{sampling-pattern} is chosen to identify important blocks. However, if the object is large -- such as a truck or bus -- 
the $5 \times 5$ block sampling pattern is selected instead. Since a larger object may span across two radar blocks (the length of a truck or a bus could be around 14 meters), it is necessary to cover two adjacent blocks around the vehicle. 
Also, while the radar data is converted from polar to Cartesian domain, the polar blocks closer to the autonomous vehicle will translate to a smaller area in the Cartesian domain while the more distant polar blocks will occupy a larger area in the Cartesian domain. 
In the case of Rad-Info-2, motivated by the above reasoning, a  T-block sampling pattern is used, where only adjacent polar blocks that are close to the autonomous vehicle are chosen while considering additional polar blocks which are far away from the vehicle. This ensures that a considerably larger area around the important object is sampled while preserving sampling rates from closer polar blocks that translate to a smaller area in the Cartesian domain. 
The T-block sampling pattern is used instead of the $5 \times 5$ block and we use the $3 \times 3$ block in case of small objects that appear within 50 meters range. In order to give importance to objects that appear very close to the AV, we also sample 3 adjacent blocks on either side of the predicted object while the object is within 16 meters of the AV in a given frame. 

Once the number of important and other polar blocks are identified, we use LP with the constraints specified below to determine the sampling rate for the important and other blocks. 
For any vector $x\in \mathbb{R}^2$, let 
\begin{align*}
    f(x) &= I*w*h*x_1 + O*w*h*x_2.
\end{align*}
Then, we have the following linear program
\begin{align*}
\max_{x\geq 0} &\; f(x)\\
s.t. &\; x_1 >= 1.1x_2\\
& f(x) \leq S, \;x_1l \leq x_1 \leq x_1u,\\
& x_2l\leq x_2\leq x_2u. \\
\end{align*}
In the above LP, $w$ is the width of 48 (range), $h$ is the height of 20 (azimuth) of the block, $I$ denotes the total number of important blocks, and $O$ is the total number of other blocks. In total, since the $400 \times 576$ frame is split into $20 \times 48$, there are 240 blocks. $x_1$ is the sampling rate for important blocks and $x_2$ is the sampling rate for the other blocks. The $f(x) < S$ condition is used to limit the number of samples -- e.g., to 10\% or 20\% or 30\% of the total samples (400x576). The condition $x_1 >= 1.1x_2$ aids in ensuring that the sampling rate for the important regions is higher than for the other regions. Finally, the lower bound for $x_1$ is 0.1, 0.2, and 0.3 for the three sampling rates 10\%, 20\%, and 30\%, respectively, while the upper bound is 0.55. The lower bounds were chosen to ensure that the sampling rate is at least as in the standard-CS case, while the upper bound was determined such that the reconstruction matches that of the original. In case of $x_2$, the lower bound was set to 0.07 to ensure there are enough samples to support reconstruction, and the upper bounds were set to 0.1, 0.2 and 0.3 for 10\%, 20\% and 30\% sampling rates, respectively, since the number of measurements could be limited due to lack of object of interest in the other regions. Once the sampling rates are determined by solving the LP, they can be used for the subsequent radar frame and the reconstructed radar is used anew for object detection; this, in turn, is further used for important region determination, and the loop continues for 20 frames. 
Therefore, in case of 10\% sampling rate, S is set to 23040 (10\% of $400 \times 576$) and these measurements are adaptively allocated across the important and other regions of the radar frame based on the LP results while maintaining the overall sampling budget to be within 10\% or 23040 measurements for 10\% sampled radar data. 

\subsection{RAD-Net}
The network, proposed in \cite{radiate}, takes a single Cartesian radar frame as input and predicts bounding boxes and classes. The RAD-Net is a modification of the Faster-RCNN \cite{fasterRCNN} network with pre-defined anchor sizes and a single class prediction head. The network is trained on both good and bad weather condition data using ResNet-50 \cite{ResNet} as a backbone. We generate the fine-tuning dataset using a pipeline similar to the one proposed in Figure \ref{Algorithm}. First, the fully sampled radar data is used as the first frame to predict important radar regions for the second frame; that information is then utilized to sub-sample the second radar frame. For the third radar frame, we rely on the original second radar frame to predict bounding boxes and important blocks, and use this information to reconstruct the frame. Proceeding in this way, 200 images were generated as the fine-tuning set.

\section{Experimental Results}
\begin{figure*}[ht!]
\begin{center}
\includegraphics[width=16.8cm]{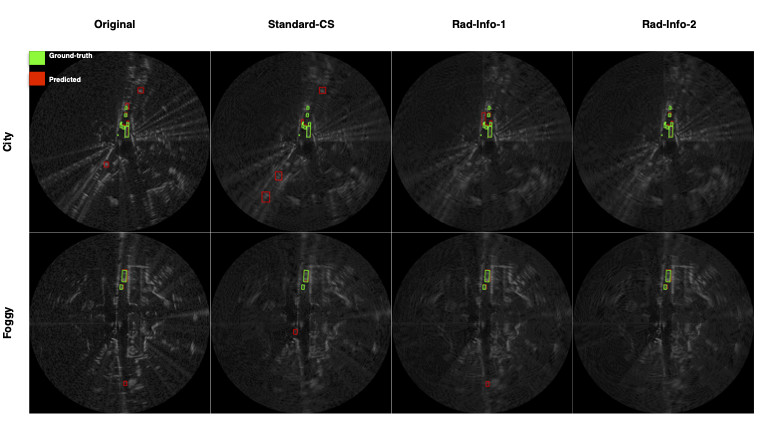}
\centering
\caption{\label{object-detection-result}Object detection on original radar, standard-CS, and on data sampled at 20\% rate and reconstructed using Rad-Info-1 and Rad-info-2.}
\label{Algo}
\end{center}
\vspace{-6mm}
\end{figure*}
We tested the above algorithm on 5 scenes: city, motorway, night and snow with 20 frames each, and foggy with 18 frames. We report the standard AP50 metric \cite{radiate}, the average precision at IoU of 0.5 and AP, the average precision calculated and averaged across IoU ranging from 0.50 to 0.95. The standard-CS resulted in a poor reconstruction quality and hence low AP50 of 6.3 while using 10\% of the samples, 34.6 for 20\% of the samples, and 47.6 for 30\% of the samples. Additional reconstruction experiments that relied on 40\% of the samples in the standard-CS setting led to 54.5 AP50 and 21.5 AP. Moreover, using 50\% of the samples yielded 56.3 AP50 and 23.1 AP, while using 55\% of the samples yielded 56.7 AP50 and 23.3 AP. Given these performances, we chose 55\% as the upper bound sampling rate for our algorithm. Across all the sampling rates shown in Table \ref{sampling}, our algorithm consistently outperforms the standard-CS algorithm with at least 10\% AP50 difference in case of 20\% and 30\% data sampling rates. For Radar-info-2, the T-block sampling pattern and assigning importance to the region closer to the AV yielded 54.2 AP50 and 22.5 AP in case of 20\% sampling rate, outperforming Radar-info-1 which achieved 47.9 AP50 and 20.9 AP. We conjecture that by saving on sampling rate in the regions adjacent to the object and thus reducing the total number of important blocks helps maintain a higher sampling rate at the remaining important blocks, leading to an improved reconstruction quality and object detection. This emphasises the need to precisely identify the region wherein an object of interest is present, and allocate most of the sampling resources to that region to ensure accurate reconstruction. In the case of the 30\% sampling rate, the reconstruction using Rad-info-2 achieves AP almost as same as that of the scheme using all radar data. 
In Figure \ref{object-detection-result}, we show detection results across the baseline and our proposed algorithm on a frame from fog and city conditions. 


\begin{table}[ht!]
\caption{\label{sampling}
Radar reconstruction results [AP/AP50]. Original radar AP50: 60.6 and AP: 23.4}
\centering
\begin{tabular}{|c| c| c| c|} 
 \hline
 Sampling rate & 10\% & 20\% & 30\% \\ [0.5ex] 
 \hline\hline
 Standard-CS &   6.3/2.4 & 34.5/13.1 &47.6/18.9\\
 \hline
 Radar Info - 1 &  29.3/12.0 & 47.9/20.9 &56.7/23.3\\
 \hline
 Radar-Info - 2 &   34.6/14.9 &54.2/22.5 &57.6/23.8\\
 \hline
\end{tabular}
\end{table}



\begin{figure}[ht!]
\begin{center}
\includegraphics[width=8cm]{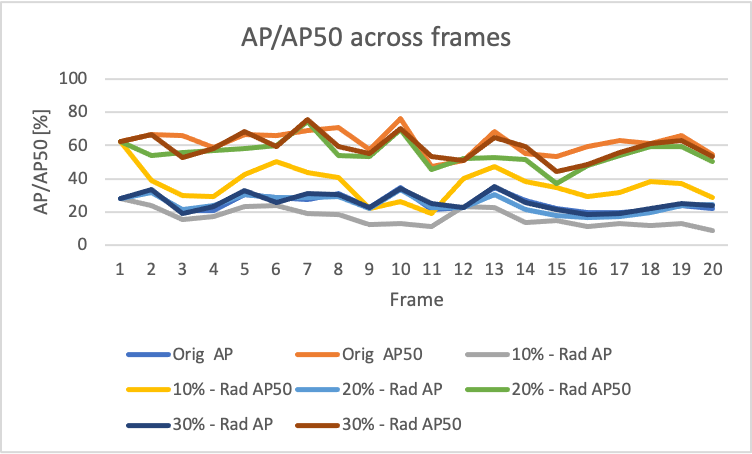}
\centering
\caption{\label{propagation}AP/AP50 evaluated per frame across scenes.}
\label{Algo}
\end{center}
\vspace{-6mm}
\end{figure}

\begin{table}[ht!]
\caption{\label{finetune}
Fine-tuning results [AP/AP50] for 20\% sampled radar data. FT: Fine-tuning}
\centering
\begin{tabular}{|c| c| c| c| c| c| c|} 
 \hline
 Train-set & Overall & city & foggy & snow & night & motorway \\ [0.5ex] 
 \hline\hline
 \multirow{2}{4em}{Before FT} & 54.2 &54.8 &72.8 &58.9 &85.0 &39.9\\
 &   22.5 &22.7 &46.0 &28.5 &47.6 &11.4\\
 \hline
 
 \multirow{2}{4em}{Original} &  53.5 &54.8 &72.7 &58.5 &81.2 &39.8\\
 &  22.2 &22.7 &46.0 &27.4 &46.6 &11.4\\
 \hline
 \multirow{2}{4em}{20\% radar} &   55.3 &55.8 &72.4 &57.9 &85.1 &42.9\\
 &   22.3 &23.0 &46.4 &26.2 &46.5 &11.9\\
 \hline
\end{tabular}
\vspace{-2mm}
\end{table}
Since only the first radar frame is fully sampled, we also analyse the AP/AP50 results averaged across different scenes for each frame to ensure no propagation error has affected the reconstruction quality of the subsequent frames. Figure \ref{propagation} shows AP and AP50 for each frame on the original, 10\%, 20\%, and 30\% sampled radar frames. The first frames yield the same AP since they are fully sampled in all cases. In the subsequent frames, at 20\% and 30\% sampling rates, the AP curve follows the performance of the original radar AP curve until the 20\textsuperscript{th} frame. However, AP performance for 10\% sampled radar data deteriorates, especially from the 7\textsuperscript{th} to the 11 \textsuperscript{th} frame. Similarly, in the case of 20\% and 30\% sampled radar frames, AP50 closely follows AP50 of the full radar data scheme, i.e., there is no evidence of propagation error. 

Finally, we fine-tune the network on the 20\% sampled reconstructed radar data. We hypothesize that tuning the network to predict objects on the sub-sampled radar data should aid in improving detection accuracy since the network learns to detect objects better in regions sampled at a higher rate than in other sub-sampled regions. The object detection results on the fine-tuned network are shown in Table \ref{finetune}. In the first row, we show the detection results of the 20\% sampled radar data reconstructed using the Rad-Info-2 algorithm. Without fine-tuning, the achieved AP50 and AP were 54.2 and 22.5, respectively. We use 200 images for fine-tuning, set the learning rate to $10^{-6}$, and fine-tune for 100 iterations, with a batch size of 2. This results in an overall AP50 of 55.3 and AP of 22.3. Although there is a slight drop in AP, the overall AP50 increases by more than 1\%; in the case of the motorway, AP50 increases by 3\% while AP also slightly improves. As an ablation study, we also fine-tune the network with the fully sampled 200 radar frames to show the benefit of fine-tuning using the 20\% sampled radar data. Fine-tuning with the original radar data decreases the overall AP50 to 53.5 and AP to 22.2, indicating rapid overfitting. 
Therefore, in the case of a populated environment with many objects such as a city or motorway, where the sampling budget is shared across many objects and a wider region in the radar frame, fine-tuning with sub-sampled data improves the detection performance. 

\section{Conclusion}
In this paper, we present an end-to-end radar acquisition and radar-based object detection pipeline. The sub-sampled radar data is processed to identify important objects under multiple extreme weather conditions such as snow and fog; the location/region of the important object is used as prior knowledge for adaptive CS in the subsequent radar frame. We tested our algorithm on the RADIATE dataset across 5 scene conditions with about 20 frames each. In our experiments, the reconstruction based on data sampled at 20\% rate shows an overall AP50 of 54.2 and AP of 22.5 detection performance, while the original data yields AP50 of 60.6 and AP of 23.4. By increasing the sampling rate to 30\% our method results in AP50 of 57.6 and AP of 23.8, i.e., it achieves the original AP while AP50 that is lower than the original by 3\%. When using the 20\% sampling rate, the achieved AP is 1\% lower than the original while AP50 is 6\% lower. However, across all sampling rates, the proposed algorithm consistently outperforms the standard-CS baseline. Finally, we generated 200 radar frames sampled at 20\% and fine-tuned the object detection system, demonstrating 1.1\% improvement in overall AP50 and 3\% AP50 gain in case of motorway scene.  

\bibliographystyle{template}
\bibliography{template.bib}

\newcommand{\comment}[1]{}
\comment{
\pagebreak

This document is a model and instructions for \LaTeX.
Please observe the conference page limits. 

\section{Ease of Use}

\subsection{Maintaining the Integrity of the Specifications}

The IEEEtran class file is used to format your paper and style the text. All margins, 
column widths, line spaces, and text fonts are prescribed; please do not 
alter them. You may note peculiarities. For example, the head margin
measures proportionately more than is customary. This measurement 
and others are deliberate, using specifications that anticipate your paper 
as one part of the entire proceedings, and not as an independent document. 
Please do not revise any of the current designations.

\section{Prepare Your Paper Before Styling}
Before you begin to format your paper, first write and save the content as a 
separate text file. Complete all content and organizational editing before 
formatting. Please note sections \ref{AA}--\ref{SCM} below for more information on 
proofreading, spelling and grammar.

Keep your text and graphic files separate until after the text has been 
formatted and styled. Do not number text heads---{\LaTeX} will do that 
for you.

\subsection{Abbreviations and Acronyms}\label{AA}
Define abbreviations and acronyms the first time they are used in the text, 
even after they have been defined in the abstract. Abbreviations such as 
IEEE, SI, MKS, CGS, ac, dc, and rms do not have to be defined. Do not use 
abbreviations in the title or heads unless they are unavoidable.

\subsection{Units}
\begin{itemize}
\item Use either SI (MKS) or CGS as primary units. (SI units are encouraged.) English units may be used as secondary units (in parentheses). An exception would be the use of English units as identifiers in trade, such as ``3.5-inch disk drive''.
\item Avoid combining SI and CGS units, such as current in amperes and magnetic field in oersteds. This often leads to confusion because equations do not balance dimensionally. If you must use mixed units, clearly state the units for each quantity that you use in an equation.
\item Do not mix complete spellings and abbreviations of units: ``Wb/m\textsuperscript{2}'' or ``webers per square meter'', not ``webers/m\textsuperscript{2}''. Spell out units when they appear in text: ``. . . a few henries'', not ``. . . a few H''.
\item Use a zero before decimal points: ``0.25'', not ``.25''. Use ``cm\textsuperscript{3}'', not ``cc''.)
\end{itemize}

\subsection{Equations}
Number equations consecutively. To make your 
equations more compact, you may use the solidus (~/~), the exp function, or 
appropriate exponents. Italicize Roman symbols for quantities and variables, 
but not Greek symbols. Use a long dash rather than a hyphen for a minus 
sign. Punctuate equations with commas or periods when they are part of a 
sentence, as in:
\begin{equation}
a+b=\gamma\label{eq}
\end{equation}

Be sure that the 
symbols in your equation have been defined before or immediately following 
the equation. Use ``\eqref{eq}'', not ``Eq.~\eqref{eq}'' or ``equation \eqref{eq}'', except at 
the beginning of a sentence: ``Equation \eqref{eq} is . . .''

\subsection{\LaTeX-Specific Advice}

Please use ``soft'' (e.g., \verb|\eqref{Eq}|) cross references instead
of ``hard'' references (e.g., \verb|(1)|). That will make it possible
to combine sections, add equations, or change the order of figures or
citations without having to go through the file line by line.

Please don't use the \verb|{eqnarray}| equation environment. Use
\verb|{align}| or \verb|{IEEEeqnarray}| instead. The \verb|{eqnarray}|
environment leaves unsightly spaces around relation symbols.

Please note that the \verb|{subequations}| environment in {\LaTeX}
will increment the main equation counter even when there are no
equation numbers displayed. If you forget that, you might write an
article in which the equation numbers skip from (17) to (20), causing
the copy editors to wonder if you've discovered a new method of
counting.

{\BibTeX} does not work by magic. It doesn't get the bibliographic
data from thin air but from .bib files. If you use {\BibTeX} to produce a
bibliography you must send the .bib files. 

{\LaTeX} can't read your mind. If you assign the same label to a
subsubsection and a table, you might find that Table I has been cross
referenced as Table IV-B3. 

{\LaTeX} does not have precognitive abilities. If you put a
\verb|\label| command before the command that updates the counter it's
supposed to be using, the label will pick up the last counter to be
cross referenced instead. In particular, a \verb|\label| command
should not go before the caption of a figure or a table.

Do not use \verb|\nonumber| inside the \verb|{array}| environment. It
will not stop equation numbers inside \verb|{array}| (there won't be
any anyway) and it might stop a wanted equation number in the
surrounding equation.

\subsection{Some Common Mistakes}\label{SCM}
\begin{itemize}
\item The word ``data'' is plural, not singular.
\item The subscript for the permeability of vacuum $\mu_{0}$, and other common scientific constants, is zero with subscript formatting, not a lowercase letter ``o''.
\item In American English, commas, semicolons, periods, question and exclamation marks are located within quotation marks only when a complete thought or name is cited, such as a title or full quotation. When quotation marks are used, instead of a bold or italic typeface, to highlight a word or phrase, punctuation should appear outside of the quotation marks. A parenthetical phrase or statement at the end of a sentence is punctuated outside of the closing parenthesis (like this). (A parenthetical sentence is punctuated within the parentheses.)
\item A graph within a graph is an ``inset'', not an ``insert''. The word alternatively is preferred to the word ``alternately'' (unless you really mean something that alternates).
\item Do not use the word ``essentially'' to mean ``approximately'' or ``effectively''.
\item In your paper title, if the words ``that uses'' can accurately replace the word ``using'', capitalize the ``u''; if not, keep using lower-cased.
\item Be aware of the different meanings of the homophones ``affect'' and ``effect'', ``complement'' and ``compliment'', ``discreet'' and ``discrete'', ``principal'' and ``principle''.
\item Do not confuse ``imply'' and ``infer''.
\item The prefix ``non'' is not a word; it should be joined to the word it modifies, usually without a hyphen.
\item There is no period after the ``et'' in the Latin abbreviation ``et al.''.
\item The abbreviation ``i.e.'' means ``that is'', and the abbreviation ``e.g.'' means ``for example''.
\end{itemize}
An excellent style manual for science writers is \cite{b7}.

\subsection{Authors and Affiliations}
\textbf{The class file is designed for, but not limited to, six authors.} A 
minimum of one author is required for all conference articles. Author names 
should be listed starting from left to right and then moving down to the 
next line. This is the author sequence that will be used in future citations 
and by indexing services. Names should not be listed in columns nor group by 
affiliation. Please keep your affiliations as succinct as possible (for 
example, do not differentiate among departments of the same organization).

\subsection{Identify the Headings}
Headings, or heads, are organizational devices that guide the reader through 
your paper. There are two types: component heads and text heads.

Component heads identify the different components of your paper and are not 
topically subordinate to each other. Examples include Acknowledgments and 
References and, for these, the correct style to use is ``Heading 5''. Use 
``figure caption'' for your Figure captions, and ``table head'' for your 
table title. Run-in heads, such as ``Abstract'', will require you to apply a 
style (in this case, italic) in addition to the style provided by the drop 
down menu to differentiate the head from the text.

Text heads organize the topics on a relational, hierarchical basis. For 
example, the paper title is the primary text head because all subsequent 
material relates and elaborates on this one topic. If there are two or more 
sub-topics, the next level head (uppercase Roman numerals) should be used 
and, conversely, if there are not at least two sub-topics, then no subheads 
should be introduced.

\subsection{Figures and Tables}
\paragraph{Positioning Figures and Tables} Place figures and tables at the top and 
bottom of columns. Avoid placing them in the middle of columns. Large 
figures and tables may span across both columns. Figure captions should be 
below the figures; table heads should appear above the tables. Insert 
figures and tables after they are cited in the text. Use the abbreviation 
``Fig.~\ref{fig}'', even at the beginning of a sentence.

\begin{table}[htbp]
\caption{Table Type Styles}
\begin{center}
\begin{tabular}{|c|c|c|c|}
\hline
\textbf{Table}&\multicolumn{3}{|c|}{\textbf{Table Column Head}} \\
\cline{2-4} 
\textbf{Head} & \textbf{\textit{Table column subhead}}& \textbf{\textit{Subhead}}& \textbf{\textit{Subhead}} \\
\hline
copy& More table copy$^{\mathrm{a}}$& &  \\
\hline
\multicolumn{4}{l}{$^{\mathrm{a}}$Sample of a Table footnote.}
\end{tabular}
\label{tab1}
\end{center}
\end{table}

\begin{figure}[htbp]
\centerline{\includegraphics{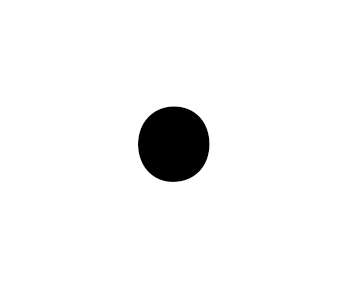}}
\caption{Example of a figure caption.}
\label{fig}
\end{figure}

Figure Labels: Use 8 point Times New Roman for Figure labels. Use words 
rather than symbols or abbreviations when writing Figure axis labels to 
avoid confusing the reader. As an example, write the quantity 
``Magnetization'', or ``Magnetization, M'', not just ``M''. If including 
units in the label, present them within parentheses. Do not label axes only 
with units. In the example, write ``Magnetization (A/m)'' or ``Magnetization 
\{A[m(1)]\}'', not just ``A/m''. Do not label axes with a ratio of 
quantities and units. For example, write ``Temperature (K)'', not 
``Temperature/K''.

\section*{Acknowledgment}

The preferred spelling of the word ``acknowledgment'' in America is without 
an ``e'' after the ``g''. Avoid the stilted expression ``one of us (R. B. 
G.) thanks $\ldots$''. Instead, try ``R. B. G. thanks$\ldots$''. Put sponsor 
acknowledgments in the unnumbered footnote on the first page.

\section*{References}

Please number citations consecutively within brackets \cite{b1}. The 
sentence punctuation follows the bracket \cite{b2}. Refer simply to the reference 
number, as in \cite{b3}---do not use ``Ref. \cite{b3}'' or ``reference \cite{b3}'' except at 
the beginning of a sentence: ``Reference \cite{b3} was the first $\ldots$''

Number footnotes separately in superscripts. Place the actual footnote at 
the bottom of the column in which it was cited. Do not put footnotes in the 
abstract or reference list. Use letters for table footnotes.

Unless there are six authors or more give all authors' names; do not use 
``et al.''. Papers that have not been published, even if they have been 
submitted for publication, should be cited as ``unpublished'' \cite{b4}. Papers 
that have been accepted for publication should be cited as ``in press'' \cite{b5}. 
Capitalize only the first word in a paper title, except for proper nouns and 
element symbols.

For papers published in translation journals, please give the English 
citation first, followed by the original foreign-language citation \cite{b6}.

\vspace{12pt}
\color{red}
IEEE conference templates contain guidance text for composing and formatting conference papers. Please ensure that all template text is removed from your conference paper prior to submission to the conference. Failure to remove the template text from your paper may result in your paper not being published.
}

\end{document}